\documentclass{article}

\usepackage{microtype}
\usepackage{graphicx}
\usepackage{subfigure}
\usepackage{booktabs} 

\usepackage{hyperref}

\usepackage[accepted]{icml2024}

\usepackage{amsmath}
\usepackage{amssymb}
\usepackage{mathtools}
\usepackage{amsthm}

\usepackage[capitalize,noabbrev]{cleveref}

\theoremstyle{plain}

\theoremstyle{definition}

\theoremstyle{remark}

\usepackage[textsize=tiny]{todonotes}
\graphicspath{ {./images/} }

\icmltitlerunning{Mechanistic Interpretability of Binary and Ternary Transformer Networks}

\begin{document}
\twocolumn[
\title{Mechanistic Interpretability of Binary and Ternary Transformers}
\author{Jason Li\\ University of Waterloo\\ \texttt{j2643li@uwaterloo.ca}}
\date{}
\maketitle








]




\begin{abstract}
Recent research \cite{wang2023bitnet, ma2024era} has proposed binary and ternary transformer networks as a way to significantly reduce memory and improve inference speed in Large Language Models (LLMs) while maintaining accuracy. In this work, we apply techniques from mechanistic interpretability to investigate whether such networks learn distinctly different or similar algorithms when compared to full-precision transformer networks. In particular, we reverse engineer the algorithms learned for the toy problem of modular addition where we find that binary and ternary networks learn similar algorithms as full precision networks. This provides evidence against the possibility of using binary and ternary networks as a more interpretable alternative in the LLM setting. 
\footnote{\label{github} Code is available at \url{https://github.com/jasonlizhengjian/MI_of_binary_transformers}}
\end{abstract}

\section{Introduction}
Binary and Ternary language models have been the topic of much attention as recent research has proposed them to be a solution for the high memory requirements demanded by modern large language models. Such networks offer great memory savings with their weights being compressed to just 1 bit or 1 ternary bit (1.58 bits) per parameter whilst still offering competitive accuracy when compared to their full-precision counterparts. 

In this work, we attempt to apply techniques from the growing field of mechanistic interpretability to gain insight into the training dynamics of such networks. Importantly, due to the discrete nature of binarized/ternarized components, we hypothesized that they may be able to learn alternative discrete algorithms which are more interpretable. We study the problem of modular addition which is discrete and has been well studied in mechanistic interpretability. We compare the algorithm learned by the binary transformer versus that of the full-precision transformer to determine whether there exist any advantages to the binary transformer in terms of interpretability. 
To the best of our knowledge, this is the first work to investigate binary or ternary networks using mechanistic interpretability. 

\section{Background}
\subsection*{Mechanistic Interpretability and Modular Addition}
Mechanistic interpretability is a field focused on interpretability through reverse engineering the algorithms learned by neural networks. Modular addition is the problem of computing $a + b \mod{p}$ for a fixed prime $p$. \cite{nanda2022progress} reverse-engineered the algorithm learned by a 1-layer transformer for modular addition. In that work, the authors used a setup which involved a 1-layer transformer with all unnecessary components removed. This algorithm has been called the "clock" algorithm by \cite{zhong2023clock}. As a brief description, the clock algorithm starts by embedding $a,b$ into rotations by a Fourier transformation. Next, it adds the two rotations together using a trigonometric identity. Lastly, it rotates backwards by the rotation corresponding to $c$ for each $c \in \{0, 1, 2, \dots, P-1 \}$ to calculate each logit. \cite{zhong2023clock} also showed that alternative algorithms (such as one which they called "pizza") can also be learned by models. Additionally, they showed that it is possible for multiple different algorithms to be represented in parallel. 

\subsection*{Binary and Ternary Transformer Networks}
Binary neural networks and Ternary neural networks \cite{courbariaux2016binarized, Qin_2020} have been an area of significant research due to their advantage in computational and memory efficiency. Generally speaking, binary neural networks are structured in the same way as standard neural networks except for the fact that the weights binarized meaning restricted to $\{-1, +1 \}$). Ternarized weights are restricted to $\{-1, 0 , +1 \}$.

\cite{wang2023bitnet, ma2024era} propose the binary transformer BitNet and the ternary counterpart BitNet b1.58. Our work primarily draws on the techniques and transformer implementation from BitNet. BitNet uses the weight matrix binarization function of: $$\text{Binarize}(W) = \text{Sign}(W - \overline{W})$$
Where $\overline{W}$ is the mean of all elements of $W$. 
BitNet uses the straight-through estimator (STE) which approximates the gradient for the binarized weights, bypassing the non-differentiable functions. BitNet b1.58 performs the ternarization of weights in a slightly more complicated manner using the function of:
$$
\text{Binarize}(W) = \text{RoundClip} \left(\frac{W}{\gamma + \epsilon}, -1, 1 \right)
$$
$$
\text{RoundClip}(x,a,b) = \max(a , \min(b, \text{round}(x)))
$$
$$
\gamma = \frac{1}{nm}\sum_{ij}|W_{ij}|
$$
BitNet and BitNet b1.58 have been shown to perform competitively against full-precision transformer LLMs of the same size. BitNet b1.58 is claimed to achieve approximately equal performance to a StableLM when both are at 3B parameters. 

There are also other works \cite{liu2022bit, he2023bivit} which propose different implementations of binary transformers. We chose to use the BitNet implementation as it was the simplest.

\section{Setup}
For the transformer, our implementation uses a similar setup as BitNet \cite{wang2023bitnet, ma2024era}. For the experimental setup, it is similar to \cite{nanda2022progress} in that we use a one-layer transformer to learn modular addition for a relatively small prime. Specifically, our transformer has the same internal dimensions as \cite{nanda2022progress} in that it uses $P = 113$, residual stream dimension of $d = 128$, 4 attention heads and $n = 512$ hidden dimension of the MLP. One notable difference (apart from the binarization) is that we had to add a RMSNorm layer before the unembed. 
\medskip

Our main experiment tested a binarized network but found very similar results with a ternary network. For the binarization, most of the linear layers of the transformer are replaced with binarized equivalents which are implemented using a binary linear layer. The embed and unembed layers were not binarized but all other layers were.  The ternary network setup simply has the binary linear layers replaced with ternary linear layers in the same way as \cite{ma2024era}.
\medskip

Training was done using the AdamW optimizer \cite{loshchilov2019decoupled}. We used the same parameters as \cite{nanda2022progress} with $\gamma = 0.001$, and $ \lambda = 1$. Weight regularization was found to be necessary, this is explained in \ref{grokking}. We trained for 10,000 epochs. The binary/ternary linear layers are optimized using the straight-through estimator technique in the same way as \cite{wang2023bitnet, ma2024era}. 

\section{Results}

\begin{figure}[h!]
    \includegraphics[width=\linewidth]{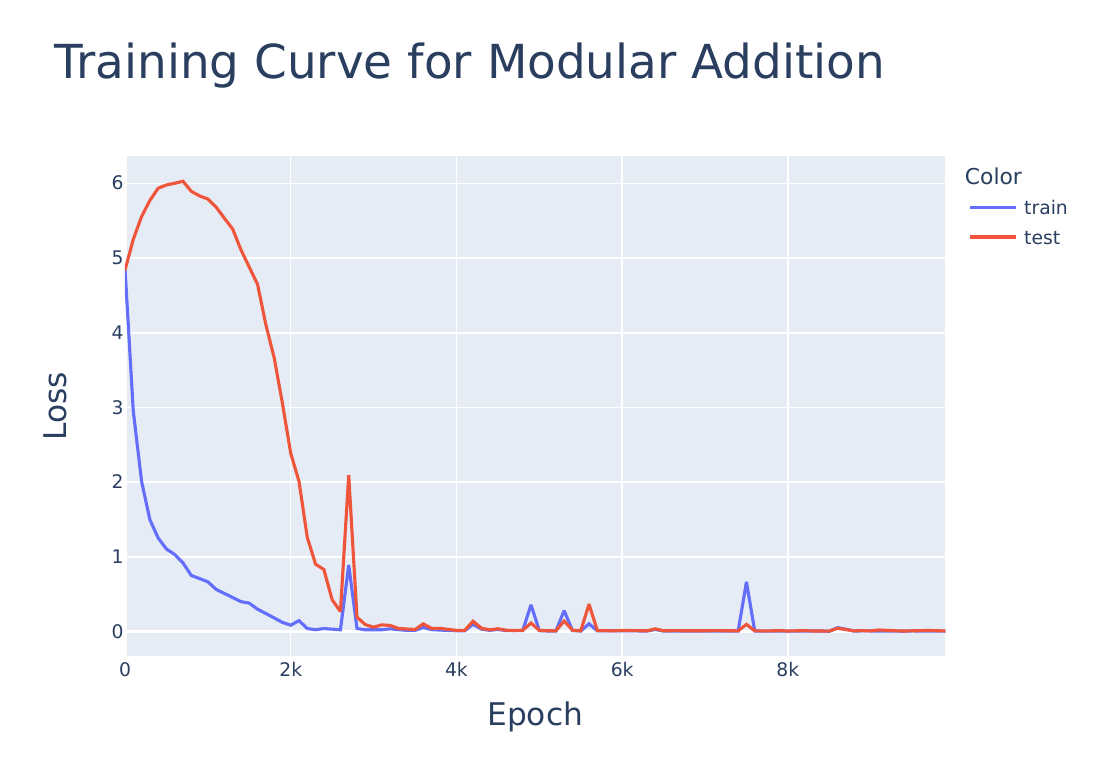}
    \caption{The curve showing training loss and test loss for 10,000 epochs. The train loss converges to around 0.006, and the test loss converges to around 0.01.}
    \label{fig:training_mod_addition}
\end{figure}

\subsection{Binary Networks and Grokking}
\label{grokking}
We find that under weight decay regularization, both binary and ternary networks consistently exhibit grokking for the tasks tested. We also find that without weight decay, binary and ternary networks do not exhibit grokking. This result behaviour matches that of full-precision networks as found in \cite{nanda2022progress}. The grokking phenomenon can be 
observed in our loss graph in figure \ref{fig:training_mod_addition}. One difference to the full precision results is that the binarized networks tend to grok in much fewer epochs as well as overfitting less in the start. One might hypothesize that this is due to the less expressive nature of binarized networks and the constraints which prevent binarized weights from exceeding 1 in magnitude. 
\begin{figure}
    \centering
    \includegraphics[width=\linewidth]{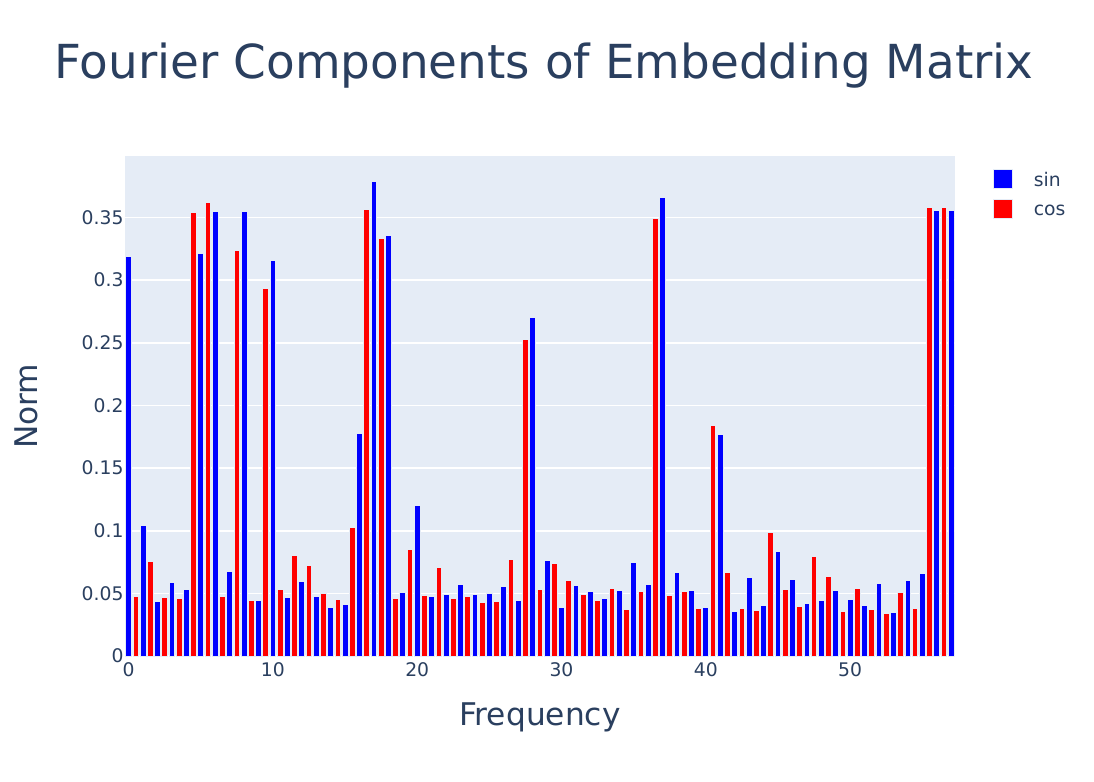}
    \caption{The norms of the Fourier components of the embedding matrix. Observe that there are a few key frequencies which stand out.}
    \label{fig:fourier-emb}
\end{figure}

\begin{figure}
    \begin{minipage}{0.5\linewidth}
        \centering
        \includegraphics[width=\linewidth]{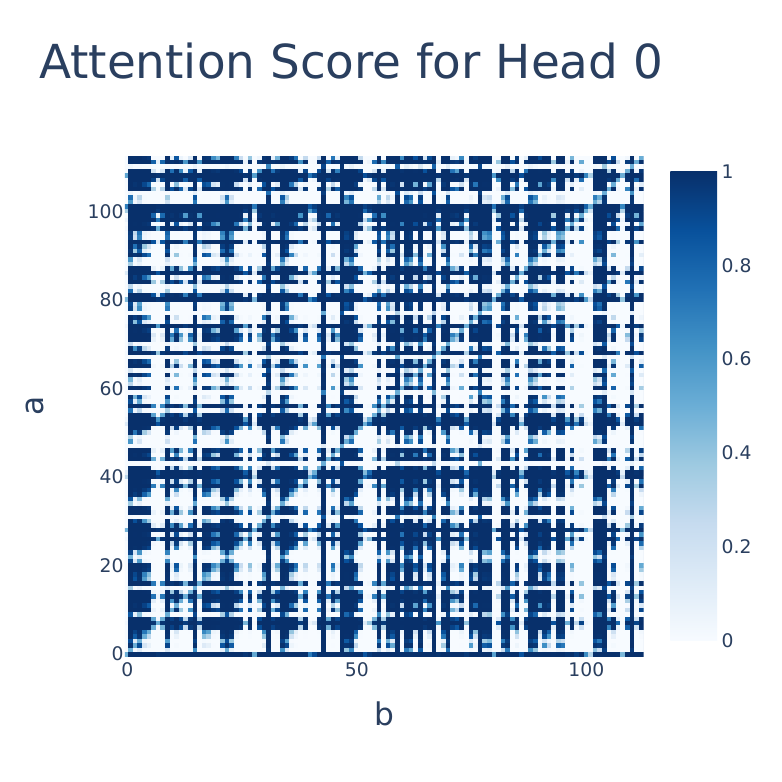}
    \end{minipage}%
    \begin{minipage}{0.5\linewidth}
        \centering
        \includegraphics[width=\linewidth]{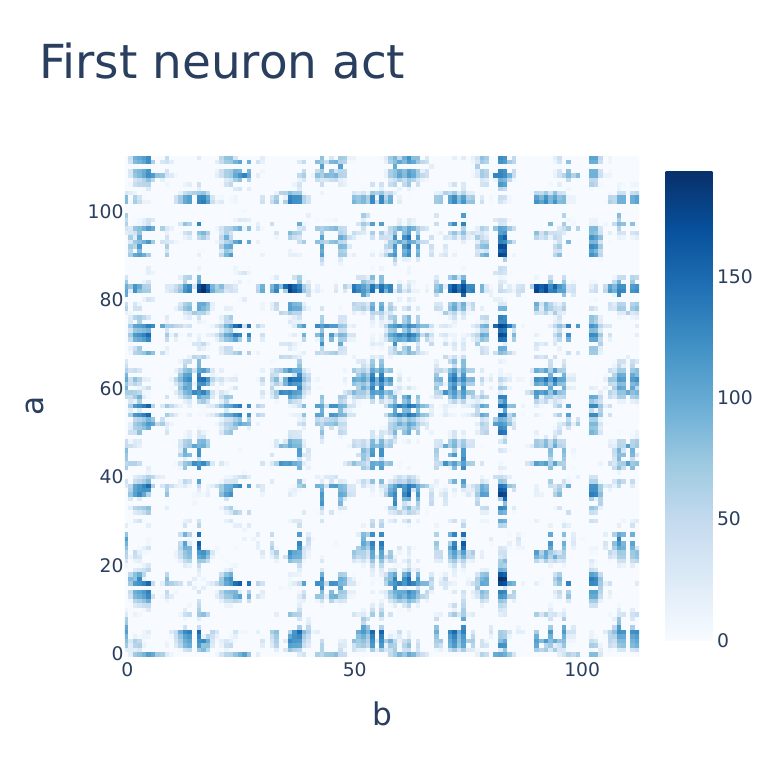}
    \end{minipage}
     \includegraphics[width=\linewidth]{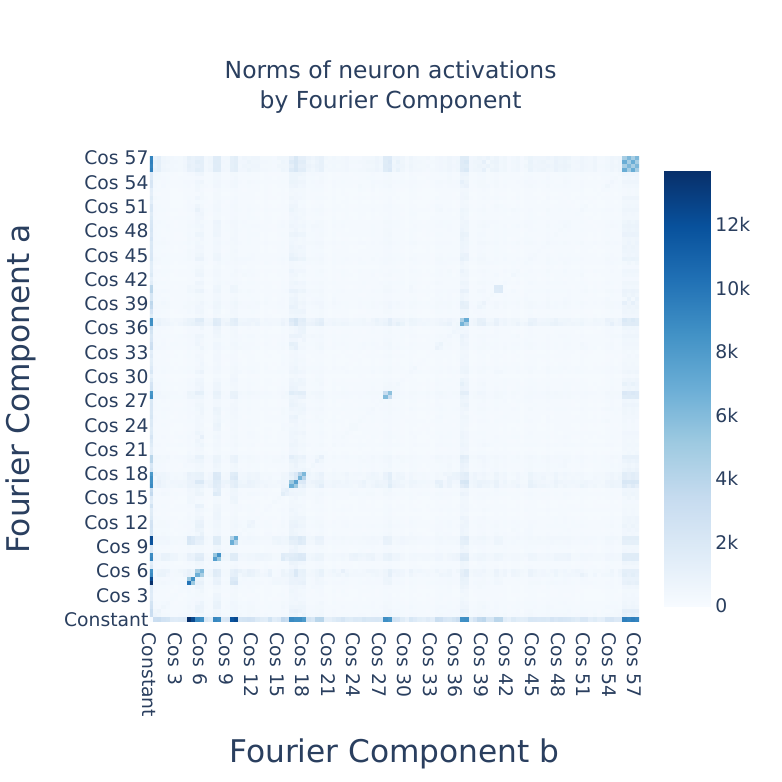}
    \caption{(Top-left) The attention score for head 0 from token '=' to token $a$ as a function of inputs $a,b$. (Top-right) The activations of MLP neuron 0. (Bottom) The norm of the Fourier components of logits. All of these correspond to the binary model.} 
    \label{fig:attention}
\end{figure}

\begin{figure}
    \centering
    \begin{minipage}{0.5\linewidth}
        \centering
        \includegraphics[width=\linewidth]{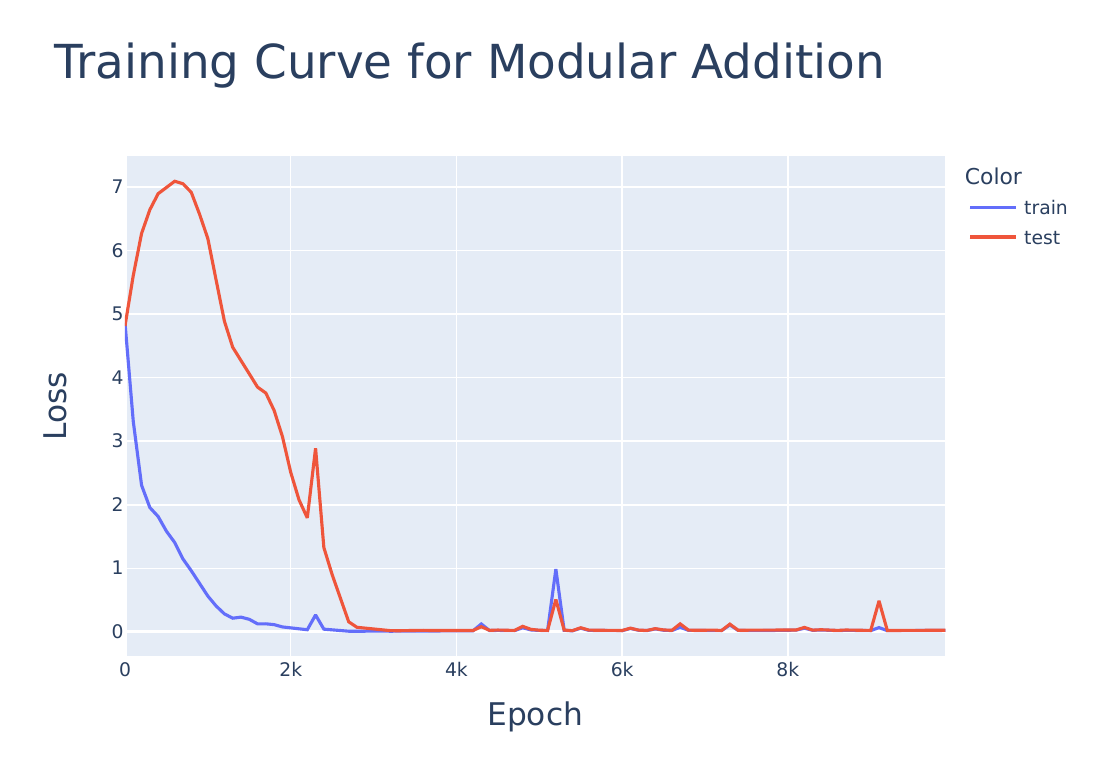}
    \end{minipage}%
    \begin{minipage}{0.5\linewidth}
        \centering
        \includegraphics[width=\linewidth]{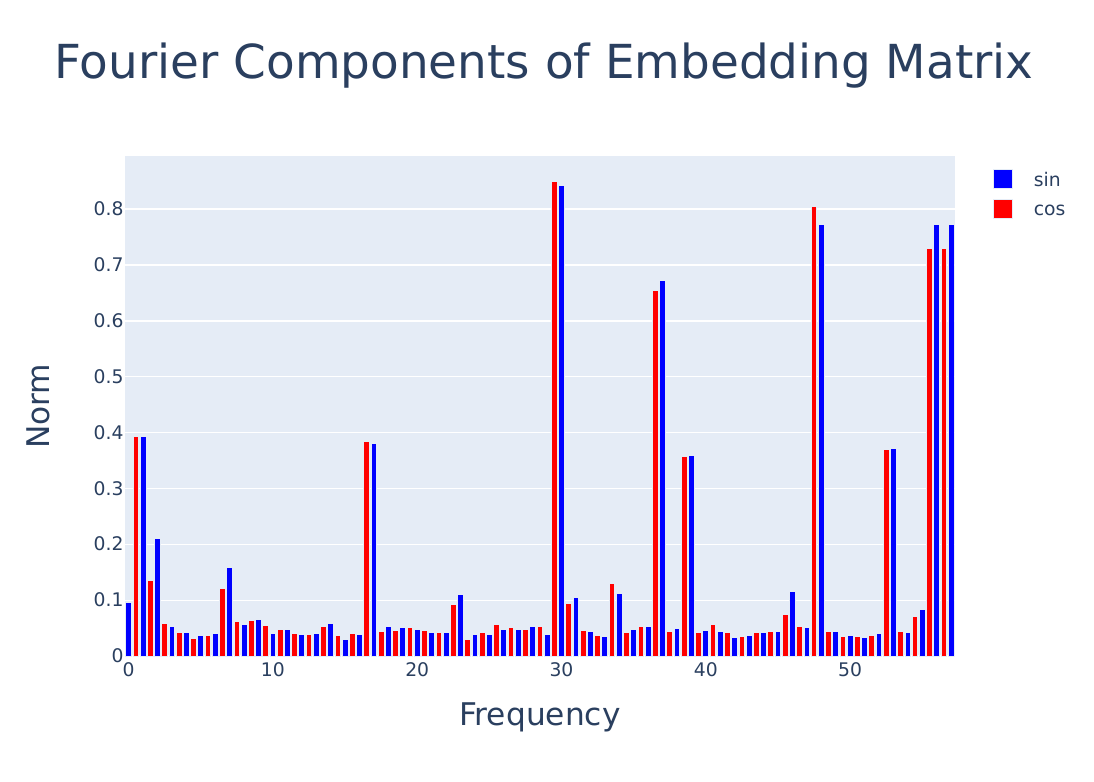}
    \end{minipage}
    \caption{For a ternary model: (Left) Loss curve. The train and test loss converge to around 0.03 for both. (Right) The norms of the Fourier components of the embedding matrix.}
    \label{fig:ternary}
\end{figure}

\subsection{Similarities}
We start with the similarities between the results for binary versus the full-precision transformer. Firstly, we observe periodicity in the embedding matrix just as in \cite{nanda2022progress} in Figure \ref{fig:fourier-emb} with a few key frequencies. Similarly, we observe the same periodicity in logits in a similar way as the full-precision model. This is shown in Figure \ref{fig:attention}.

\begin{figure}
    \centering
    \includegraphics[width=\linewidth]{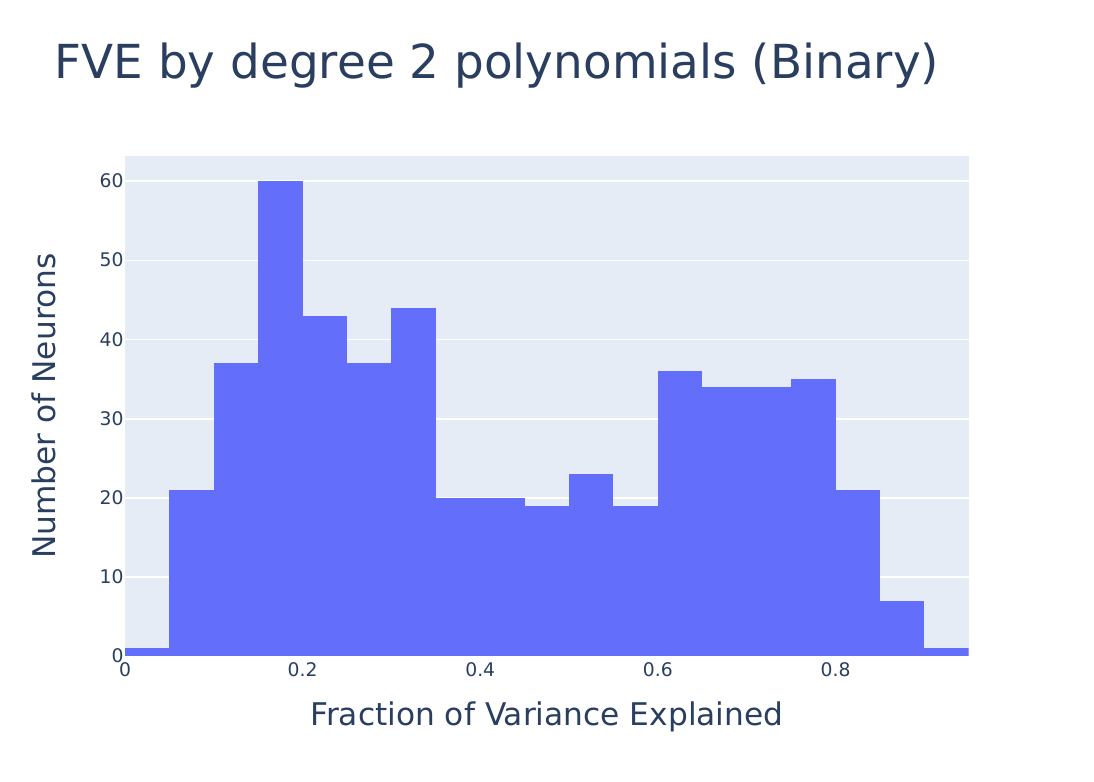}
    \includegraphics[width=\linewidth]{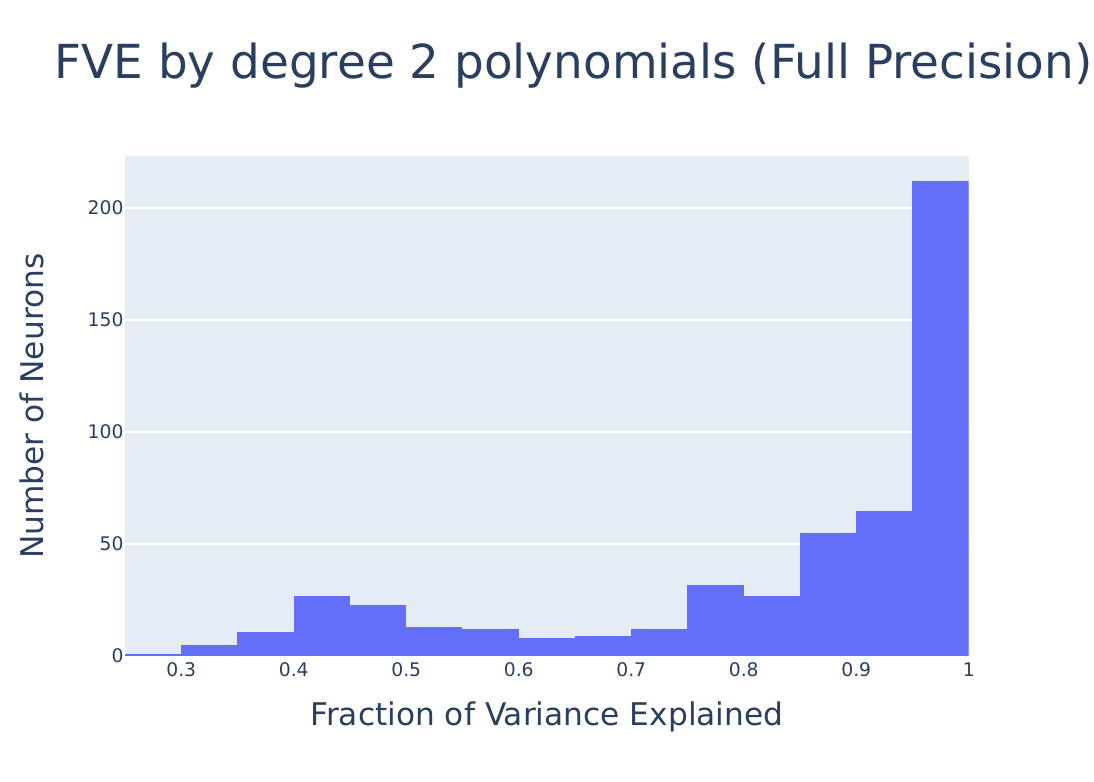}
    \caption{The fraction of variance explained by degree-2 polynomials of a single frequency. }
    \label{fig:fve}
\end{figure}

\subsection{Differences}
Next, we highlight some notable observed differences. Firstly, we observe significant noise in the projection to fourier basis in figure \ref{fig:fourier-emb} and figure \ref{fig:attention}. In Figure \ref{fig:fve} we can see a clear difference in how well degree-2 polynomials of a single frequency are capable of explaining the variance of the neurons. 

\subsection{Ternary}
\begin{figure}
    \centering
    \begin{minipage}{0.5\linewidth}
        \centering
        \includegraphics[width=\linewidth]{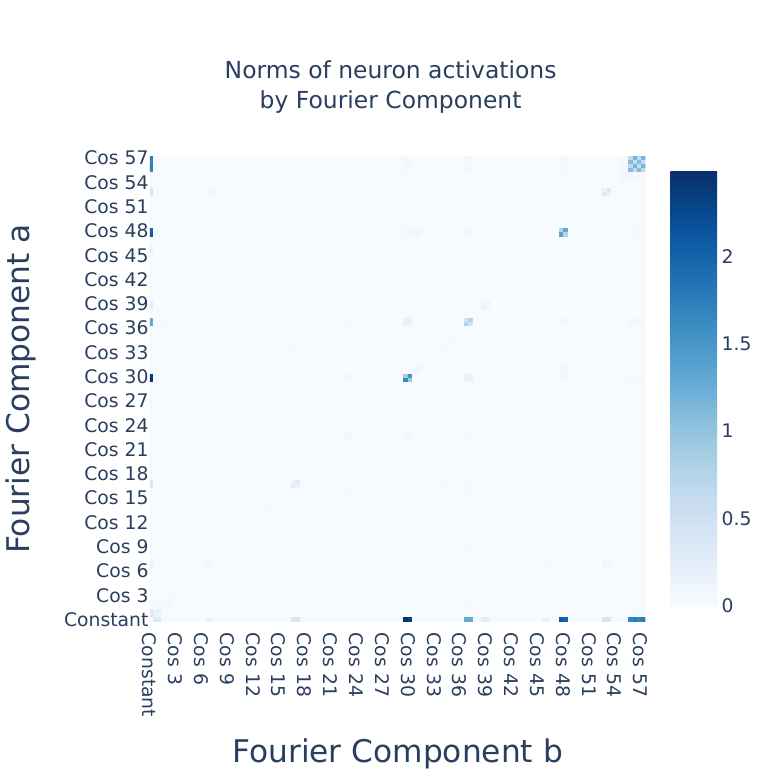}
    \end{minipage}%
    \begin{minipage}{0.5\linewidth}
        \centering
        \includegraphics[width=\linewidth]{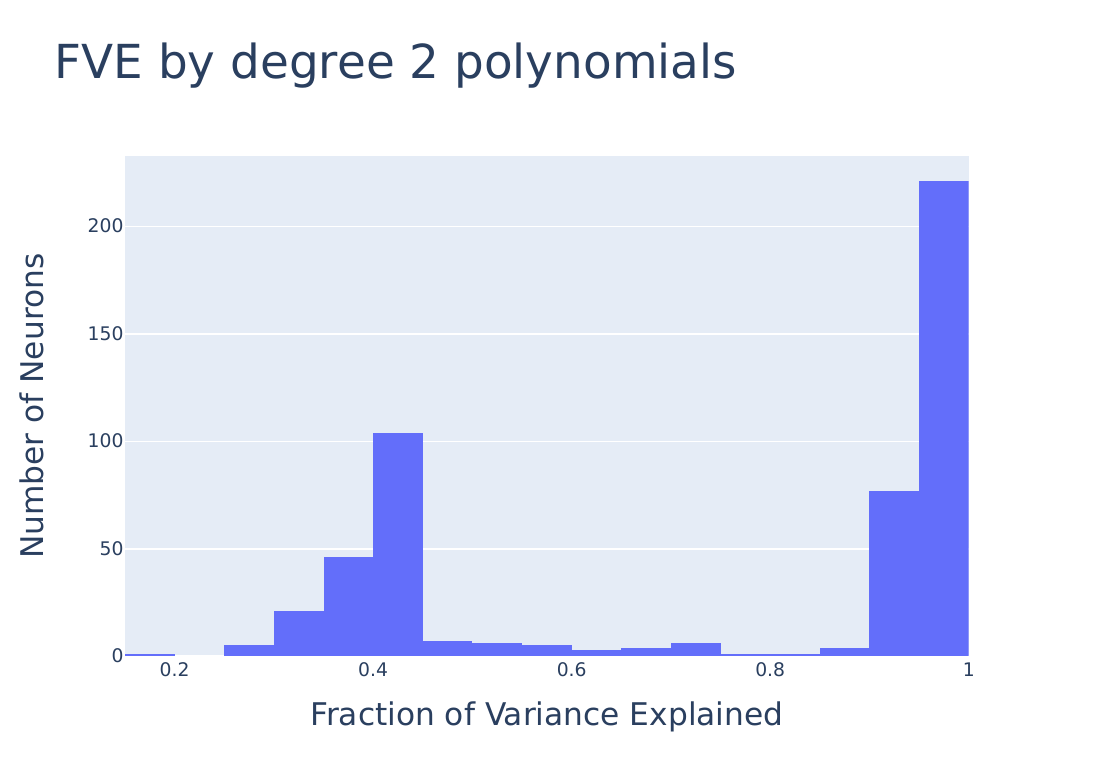}
    \end{minipage}
    \caption{For a ternary model: (Left) Norms of Fourier components of logits. (Right) The fraction of variance explained by degree-2 polynomials of a single frequency.}
    \label{fig:ternary_fve}
\end{figure}

For the ternary transformer, we observed that results fell somewhere in between the binary and the full-precision. The same similarities still held while the differences were less prevalent. Ternary models exhibit grokking, and demonstrate the same periodicity as the binary model and the full-precision model as shown by figure \ref{fig:ternary}. The ternary model exhibits less noise than the binary model (more similar to the full-precision) as shown by figure \ref{fig:ternary_fve}. An interesting note is that while the ternary model seems to behave closer to the full-precision model, it does not achieve a superior loss when compared to the binary model. 

\section{Discussion and Conclusion}
Our experiment has demonstrated that for the toy problem of modular addition, binary and ternary transformer networks exhibit very similar grokking behaviour as full precision transformers. They also tend to learn similar fourier transformation based algorithms which are not more easily interpreted when compared to the full-precision model. In fact, they appeared to be less interpretable in our analysis due to the added noise when the same techniques were applied. One limitation of our study in this regard is the lack of conclusive evidence concerning exactly what algorithm was learned, since \cite{zhong2023clock} showed that many different algorithms could be learned for the modular addition problem.

We originally hypothesized that it might be possible to learn more interpretable algorithms due to the discretized nature of binary/ternary weights. We thought that if alternative discrete algorithms could be learned, then since modular arithmetic has simple discrete algorithmic solutions, the learned algorithm may be more human-interpretable compared to the clock algorithm or other similar Fourier transform-based algorithms. The results found in this work have demonstrated evidence against the possibility of using binary and ternary to learn simpler discrete algorithms. 

In this work, we only explored the most basic straight-through estimator technique of optimizing binary/ternary networks. In practice, there are many tricks/techniques for optimizing binary and ternary networks \cite{Qin_2020}. One of these more complicated techniques is used to implement a binary transformer in \cite{liu2022bit}. Further work could also be done to explore these using mechanistic interpretability. Additionally, it is important to note that our implementation did not binarize/ternarize then embed/unembed layers of the transformer (the same as BitNet). This was because we were unable to get it to optimize when embed/unembed layers were binarized. Further work should be done to investigate networks which are fully binarized/ternarized in all layers. 

Our investigation of the specific algorithms learned by the models was limited to only the toy problem of modular addition. Further work may be done to investigate whether binary and ternary networks always tend to learn similar algorithms to full-precision counterparts for other toy problems, and to try and find cases where the binary/ternary network achieves comparable accuracy but learns a more interpretable algorithm. Future work may be done to investigate binary/ternary transformer language models in comparison to full-precision counterparts. For example, seeing if results from \cite{olsson2022context} still hold for binary/ternary networks.

Full-precision networks can represent binary/ternary networks by simply using the same weights. The reverse is obviously not true. For this reason, we believe it to be unlikely that binary/ternary networks may learn better algorithms versus a full-precision network. Also, since techniques for optimizing binary/ternary networks are still gradient-based, it is unlikely that such a network would be capable of learning a fundamentally different discrete algorithm.

\nocite{Gomez}
\bibliography{example_paper}
\bibliographystyle{icml2024}




\end{document}